\documentclass[10pt,twocolumn,letterpaper]{article}

\usepackage{cvpr}              

\usepackage[dvipsnames]{xcolor}

\usepackage{multirow}

\definecolor{cvprblue}{rgb}{0.21,0.49,0.74}
\usepackage[pagebackref,breaklinks,colorlinks,citecolor=cvprblue]{hyperref}

\def\paperID{6045} 
\def\confName{CVPR}
\def\confYear{2024}

\title{DROP: Decouple Re-Identification and Human Parsing with Task-specific Features for Occluded Person Re-identification}
\author{Shuguang Dou\\
Tongji University\\
\and
Xiangyang Jiang\\
MSRA\\
\and
Yuanpeng Tu\\
University of HongKong\\
\and
Junyao gao\\
Tongji University\\
\and 
Zefan Qu\\
Tongji University\\
\and 
Qingsong Zhao \\
Tongji University\\
\and 
Cairong Zhao\\
Tongji University\\
}

\begin{document}
\maketitle
\begin{abstract}
This paper proposes a Decouple Re-identificatiOn and human Parsing (DROP) method to learn the task-specific features that fit the two tasks for occluded person re-identification (ReID).
Currently, mainstream approaches use multi-task learning to allow for simultaneous learning of both ReID and human parsing tasks based on global features or utilize semantic information to guide attention, with the latter usually performing better. 
The paper posits that the reason for the former's inferior performance compared to the latter lies in the fact that ReID and human parsing demand features of distinct granularity. 
ReID focuses on the difference between different pedestrian parts, i.e.,~\textbf{instance part-level difference}, while human parsing focuses on the internal structure of the human body, i.e., \textbf{semantic spatial context}.
To address this, we decouple the features for person ReID and human parsing. More precisely, we propose detail-preserving upsampling to combine feature maps of varying resolutions from the backbone, decoupling the parsing-specific features for human parsing.
To further decouple the two tasks, we only add human position information to the human parsing branch to help the model learn the semantic spatial context, while in the ReID branch, we introduce the part-aware compactness loss to enhance the instance-level part difference.
Experimental results underscore the efficacy of DROP compared to the two prevailing mainstream methods, especially the Rank-1 reached 76.8\% on Occluded-Duke. The dataset and codebase of DROP are available at \href{https://github.com/shuguang-52/DROP}{https://github.com/shuguang-52/DROP}.
\end{abstract}    
\section{Introduction}
\label{sec:intro}
\begin{figure}[!t]
\centering
\includegraphics[width=\linewidth]{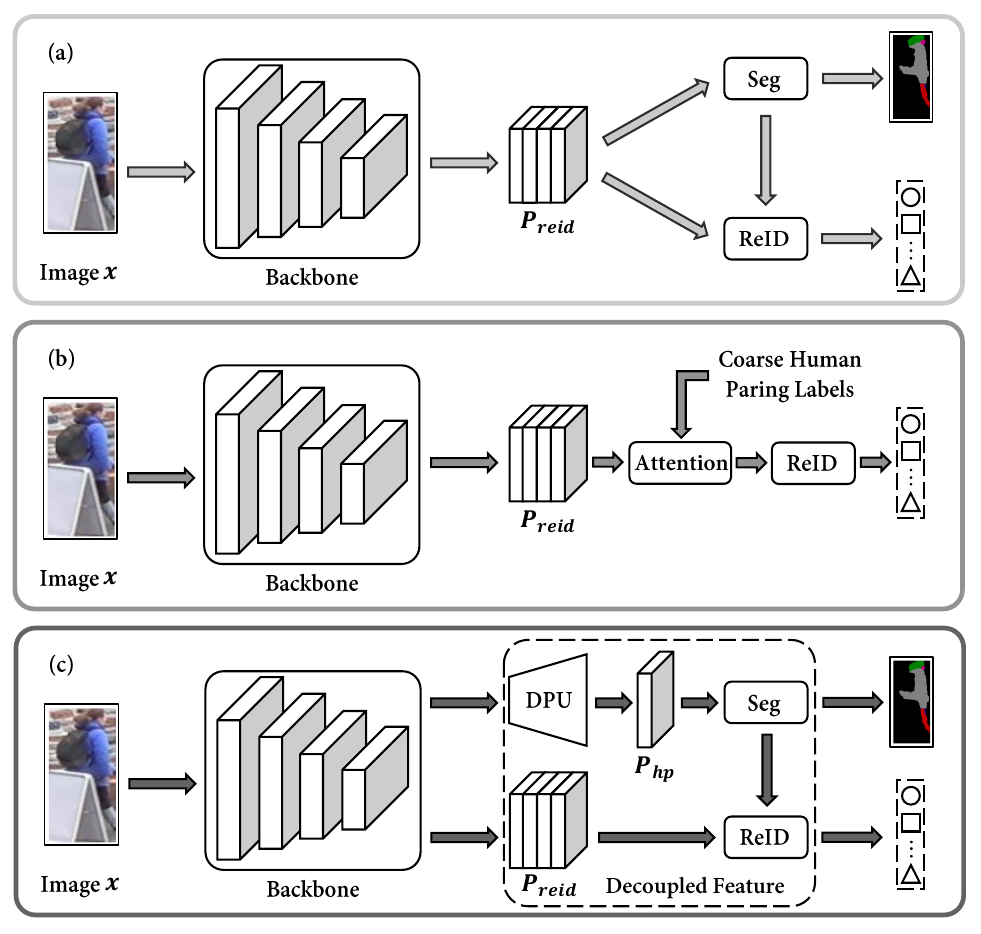}
\caption{Comparison of three methods for occluded person ReID. (a) A multi-task learning framework to simultaneously ReID and segmentation tasks based on the same features. (b) Dual Supervised attention mechanism module learning with ID labels and extra coarse human parsing labels. (c) Ours DROP.}
\label{diff}
\vspace{-3mm}
\end{figure}

Person re-identification \cite{pr_book, dl_pr} (ReID) aims to match a target pedestrian with non-overlapping cameras. However, the previous ReID approach severely degrades performance in occluded scenes due to the introduction of a large amount of noise directly matching the two occluded pedestrian images.
To solve the occlusion problem, various methods have been proposed, which can be roughly divided into Semantic information guide alignment-based methods~\cite{kepoint-pvpm, keypoint-honet, keypoint-pgfa, pixellevel-hsp, pixellevel-maskg, pixellevel-isp}, attention-based methods~\cite{attention-RGA-SC, attention-scsn, attention1, attention2, attention3, attetion-DuATM, wang2022attentive}, and data augmentation-based methods~\cite{IGOAS, FCFormer, caao}.
Although the above methods have made good progress in solving the occlusion problem from different ideas, the performance of ReID in the occluded ReID dataset still has a large performance gap with that of the Holistic ReID dataset.
The current mainstream approaches have two ways to solve the occlusion problem as shown in Fig.~\ref{sec:intro} (a) and (b). The first approach trains a multi-task learning framework to simultaneously learn person ReID and segmentation tasks based on image features from the backbone. For example, both ISP~\cite{pixellevel-isp} and HCGA~\cite{hcga} utilize two decoupled heads for both ReID and segmentation tasks based on the high-resolution feature maps generated by HRNet~\cite{HRNet}. Although the high-resolution feature maps are favorable for segmentation, the features required during the learning process of the two tasks are different and may even be conflicting. 
The second approach explores combining segmentation with attention, not directly allowing the model to learn segmentation, but rather allowing coarse human parsing labels to guide the learning of the attention mechanism thereby allowing it to focus on pedestrians. For example, SAP~\cite{SAP} encourages the attention-based partition of the (transformer) student to be partially consistent with the semantic-based teacher partition through knowledge distillation. Currently, the second approach usually achieves better results.
\par
In this paper, we explore~\emph{why multi-task learning frameworks underperform for person ReID}. 
The human parsing task focuses on localizing and classifying different body parts at the pixel level, \emph{requiring the semantic spatial context information}. Whereas the ReID task requires the model to be able to recognize nuances in pedestrians, which \emph{require attention to instance part-level difference}.
\par
To solve the above conflict issue, we explore decoupling the two tasks ReID and human parsing by learning task-specific features as shown in Fig.~\ref{sec:intro} (c). Specifically, from the backbone network, we decouple two features suitable for two different task requirements. For the human parsing task, we introduce detail-preserving upsampling (DPU) to fuse features of different depths in the backbone to obtain a high-resolution low-channel feature map. For the ReID task, in the same way as before, we directly use the low-resolution high-channel feature map output from the backbone. 
To further decouple the two tasks,  we exploit the pedestrian position encoder (PPE) to learn pedestrian position embedding from one-dimensional height coordinates. 
Since the ReID task does not require spatial context information, we only sum this embedding with the feature used for human parsing to obtain pedestrian position-aware features.
On the other hand, our method DROP introduces a memory bank to store the human parts embeddings obtained by combining the parsing results with the ReID features and proposes the part-aware compactness triplet (PCT) loss to increase the instance part-level difference by more negative samples.
During multi-task learning, we give higher learning weight to the human parsing loss different from the previous methods. Since we decouple the two tasks, the human parsing branch is better optimized with higher fitting performance without affecting the learning of ReID.
\par
We summarize the main contributions of our work as follows:
\begin{itemize}
\item We discover the inherent conflict between ReID and human parsing tasks. Instead of learning the two tasks of ReID and human parsing together, we are the first method to decouple the two tasks by learning task-specific features to the needs of the two tasks.

\item To further decouple the two tasks, we introduce a pedestrian position encoder to the human parsing branch alone to obtain pedestrian position-aware features, which is information of less interest to the ReID task.

\item We propose part-aware compactness triplet (PCT) loss to train the part-based ReID method. PCT loss exhibits robustness against occlusions and non-discriminative local appearances, making it readily integrable into various part-based frameworks.

\item Our DROP outperforms state-of-the-art methods by archiving 63.3\% mAP and 76.8\% rank-1  on the  Occluded-Duke dataset.  Our decouple method encourages further research on multi-task learning-based ReID methods.  
\end{itemize}
\section{Related work}

\paragraph{Occluded Person Re-identification.}
In real-world scenes, occlusion frequently transpires, obscuring the intended pedestrian target amid unrelated individuals within crowded environments. Zhuo~\textit{et al.}~\cite{AFPB}  pioneered the occluded person ReID challenge and introduced the Attention Framework of Person Body (AFPB) to confront this challenge.

To address the various challenges posed by occlusion, the mainstream approaches are categorized into the following three types:
\textit{a) Attention-based methods:} Those methods~\cite{attention-RGA-SC, attention-scsn, attention1, attention2, attention3, attetion-DuATM, wang2022attentive} rely on attention mechanisms to adaptively learn local discriminative features solely from ID labels.
\textit{b) Semantic information guide alignment-based methods:} Pose estimation and human parsing have been introduced to tackle occluded ReID challenges. 
Miao~\textit{et al.}~\cite{PGFA} utilize a pose estimation model to extract valuable information from occluded images, directing attention to non-occluded areas. 
Gao~\textit{et al.}~\cite{kepoint-pvpm} introduce a method for pose-guided matching of visible parts, enabling the fusion of local features with visual scores.
Wang~\textit{et al.}~\cite{keypoint-honet} initially extract semantic local features using a pose estimation model. They propose adaptive direction graph convolution layers to learn relations and a cross-graph embedded-alignment layer to predict similarity scores.
\textit{c) Data augmentation-based methods:} Several studies have suggested employing image occlusion augmentation to tackle occluded ReID challenges. This approach involves masking specific sub-regions within pedestrian images. 
Zhao~\textit{et al.}~\cite{IGOAS} introduce the Incremental Generation of Occlusion Against Suppression (IGOAS) network, generating occlusion data of varying complexity through the incremental generation of occlusion blocks. Wang~\textit{et al.}~\cite{FCFormer} present the Feature Completion Transformer (FCFormer), incorporating an Occlusion Instance Augmentation strategy to enhance the diversity of occluded training image pairs. The Content-Adaptive Auto-Occlusion (CAAO) network  integrates reinforcement learning into an automatic occlusion control module, offering adaptability to state and content, distinguishing it from previous occlusion strategies~\cite{caao}.

Recent studies have unveiled that incorporating coarse human parsing outcomes to steer attention mechanisms, particularly through the integration of these two methods, can yield more exhaustive pedestrian attention maps. The Semi-Attention Partition (SAP)~\cite{SAP} method delves into the potential of a "weak" semantic partition to effectively guide a "strong" attention-based partition.
Additionally, BPBreID ~\cite{bpbreid} introduces a soft attention mechanism trained under dual supervision, enabling the utilization of both identity and prior human parsing information.

However, the majority of occluded ReID methods simultaneously learn the ReID and semantic segmentation tasks utilizing identical image features. In this study, we introduce a decoupled approach aimed at acquiring task-specific features.
\vspace{-3mm}
\paragraph{Decoupled Heads for Multi-Task Learning.}
Object detection constitutes a classical multi-task learning paradigm wherein the model must adeptly acquire both localization and classification capabilities. Historically, the use of decoupled heads has been the prevalent setup in one-stage detectors~\cite{generalized_focal_loss, fcos, atss}. Double-Head R-CNN~\cite{DoubleHead} and TSD~\cite{TSD} reexamine the specialized sibling head extensively employed within the R-CNN family, ultimately unraveling the fundamental misalignment between classification and localization tasks. Despite highlighting the significance of decoupling these tasks, existing studies emphasize that solely decoupling at the parameter level results in an imperfect trade-off between the two tasks~\cite{TSCODE}.


\section{Method}

\begin{figure*}[!t]
\centering
\includegraphics[width=\linewidth]{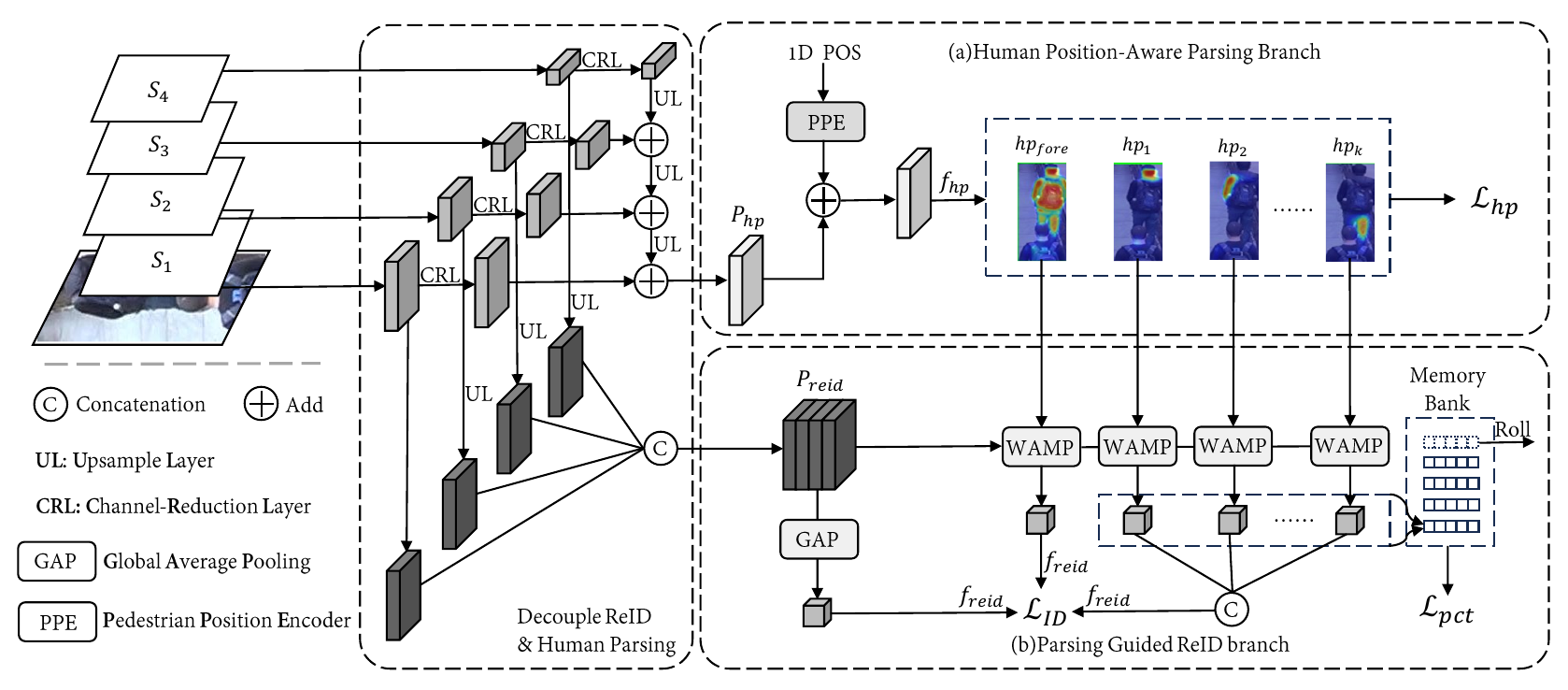}
\caption{Structure of DROP with decoupled branches. The model consists of a \textit{human position-aware parsing branch} for human parsing and a \textit{parsing guided ReID branch} for producing the global, foreground, and parts embeddings. \textit{WAMP} denotes the global weighted average and max pooling. $\mathcal{L}_{pct}$ denotes the part-aware compactness triplet loss.}
\label{fig:DROP}
\vspace{-3mm}
\end{figure*}

\subsection{Overview}
\paragraph{Motivation.}
Re-identification and human parsing represent interrelated yet contradictory tasks within occluded person ReID. ReID necessitates robust and compact features for each pedestrian, thereby demanding fine-grained details, whereas human parsing relies on coarse-grained information but requires more details and semantic spatial context. To address this divergence, mainstream methods~\cite{pixellevel-isp, hcga, bpbreid} employ decoupled heads to manage this conflict. Specifically, utilizing the final feature map $P$ obtained from the backbone, along with the ID label $y$ and coarse human parsing label $\mathcal{H}$, the model minimizes both the ID and human parsing losses independently, utilizing the same feature map $P$:

\begin{equation}
\mathcal{L}=\mathcal{L}_{reid}\left(\mathcal{F}_r\left(P, hp\right), y\right)+\lambda \mathcal{L}_{hp}\left(\mathcal{F}_p\left(P\right), \mathcal{H}\right),
\label{eq:couple}
\end{equation}
where $\mathcal{F}_r(\cdot)=\left\{f_{reid}(\cdot), \mathcal{G}(\cdot)\right\}$ and $\mathcal{F}_p(\cdot)=\left\{f_{hp}(\cdot)\right\}$ are the ReID and human parsing branches. $f_{reid}(\cdot)$ and $f_{hp}(\cdot)$ are the feature projection functions for ReID and human parsing, $hp$ is the predicted results of the human parsing branch, while $\mathcal{G}(\cdot)$ is the guide module that exploit the predicted segmentation results to get human parts embeddings. Traditionally, $f_{reid}(\cdot)$ and $f_{hp}(\cdot)$ are trained using distinct parameters to offer diverse feature contexts for each task—a configuration known as "parameter decoupling" ~\cite{TSCODE}. However, this simplistic approach falls short of fully addressing the issue, as the shared input feature map $P$ predominantly determines the semantic context. The limitation arises from the shared derivation of context, affecting its efficacy. Consequently, the conflict between ReID and human parsing induces conflicting context preferences within $P$, resulting in an imperfect equilibrium between the two tasks.

To address the issue, our proposed DROP method decouples the feature encoding for the two tasks at the source, utilizing distinct feature maps with varied semantic contexts in each branch. Departing from utilizing a shared input feature map $P$, our approach involves feeding task-specific input features, denoted as $P_{reid}$ and $P_{hp}$, into the respective branches. In pursuit of this goal, Eq. (\ref{eq:couple}) can be written as:
\begin{equation}
\mathcal{L}=\mathcal{L}_{reid}\left(\mathcal{F}_r\left(P_{reid}, hp\right), y\right)+\lambda \mathcal{L}_{hp}\left(\mathcal{F}_p\left(P_{hp}\right), \mathcal{H}\right).
\label{eq:couple}
\end{equation}
\vspace{-3mm}
\paragraph{Overall framework.}
For the ReID branch, we generate spatially coarser but semantically richer feature maps. For the human position-aware parsing branch, we provide it with feature maps containing more detailed texture and position information.
As depicted in Fig.~\ref{fig:DROP}, our approach adheres to the prevalent multi-task learning framework, comprising the backbone, the ReID branch, and the human parsing branch. The backbone produces multi-scale feature maps derived from the input images. Subsequently, our DROP branches process four levels of feature maps to produce separate feature maps for ReID and human parsing tasks.


\subsection{Human Position-aware Parsing Branch}
Unlike the ReID task, human parsing involves a more coarse-grained analysis relying on intricate texture details and semantic information to categorize pixels. However, prevailing methods typically segment ReID images from the single-scale feature map $P$. Lower-level feature maps exhibit heightened sensitivity to pedestrian contours, edges, and detailed textures, offering potential advantages for the human parsing task. Nonetheless, this advantage often incurs significant computational overhead. Methods like ISP~\cite{pixellevel-isp}, HCGA~\cite{hcga}, and BPBReID~\cite{bpbreid} integrate fused multi-scale features within HRNet to mitigate this challenge. Despite this effort, employing the same feature map for two conflicting tasks poses a significant challenge, particularly as human parsing remains an auxiliary task, hampering its optimal learning.

\paragraph{Detail-preserving upsampling.} Based on our observations, we introduce the Detail-Preserving Upsampling (DPU) method to disentangle features from the backbone, enhancing accurate parsing. DPU integrates feature maps from four stages, and its architectural depiction is illustrated in Fig.~\ref{fig:DROP} (a). For computational efficiency, we initially employ channel-reduction layers to harmonize high-channel feature maps across stages, reducing them to a uniform low channel count. Subsequently, excluding the first stage $S_1$, a 2-fold linear interpolation is employed for upsampling feature maps from $(S_2, S_3, S_4)$. To preserve detailed information richness within each stage, feature maps from deeper stages are meticulously fused with those from lower stages, introducing minimal additional parameters. Ultimately, the final feature maps of the last three stages are summed with those of the initial layer to produce the conclusive output.$P_{hp}$:
\begin{equation}
P_{hp}= \sum_{i=2}^{l-1}\operatorname{UP}\left(\operatorname{CR}\left(P_i\right)\right) + \operatorname{CR}\left(P_1\right),
\end{equation}
where $\operatorname{UP}(\cdot)$ is the upsample layer, $\operatorname{CR}(\cdot)$ is the channel-reduction layer and $i$ is the layer number of feature maps. 
\vspace{-3mm}
\paragraph{Pedestrian position-aware feature.}
Within the ReID dataset, a global spatial correlation exists between the target pedestrian's part and height. For example, the head typically appears at the top and the feet at the bottom. To facilitate the model in learning this semantic spatial context, we incorporate one-dimensional height coordinates as additional inputs to the network. 
We designed a simple pedestrian position encoding (PPE) consisting of two convolutional layers with the structure Conv-BN-ReLU-Conv-BN.
First, the 1D coordinates are expanded to the same size as the $P_{hp}$, and then the PPE is used to extract pedestrian position embedding from them. Subsequently, the embedding is added to the $P_{hp}$ output obtained from the DPU, resulting in the derivation of pedestrian position-aware features.
Finally, $f_{hp}$ outputs the results of human parsing  $\{hp_1, \cdots, hp_k\}$. We combine the predictions of each part using the max function to get the foreground mask $hp_{fore}$.
\begin{equation}
    \{hp_{fore}, hp_1, \cdots, hp_k\}=f_{hp}(P_{hp}+\operatorname{PPE}(Pos_{1D})),
\end{equation}
where $k$ denotes the number of human parts.

\subsection{Parsing Guided ReID Branch}
\label{sec:reidBranch}
This segment leverages human parsing predictions to steer ReID branch learning. We introduce the Weighted Average and Max Pooling (WAMP) technique, aggregating ReID features and parsing outcomes to derive foreground and human parts embeddings. Additionally, a parts embedding memory bank (PEMB) undergoes continuous updates during training. Leveraging this, we compute the part-aware compactness triplet loss, enhancing the robustness and compactness.
\vspace{-3mm}
\paragraph{Training.}
Figure~\ref{fig:DROP} (b) illustrates the upsampling of feature maps within the backbone, excluding the initial stage, to a uniform scale, followed by concatenation to yield $P_{reid}$. We utilize global average pooling (GAP) on $P_{reid}$ to derive the global embedding. To effectively amalgamate $P_{reid}$ and parsing results ${hp_{fore}, hp_1, \cdots, hp_k}$, we introduce a combination technique called Weighted Average and Max Pooling (WAMP), which integrates global weighted average pooling~\cite{GWAP} with maximum pooling, generating both foreground and parts embeddings. 
During the training phase, we establish a parts embedding memory bank (PEMB) sized as $[M\times B, K, C]$, where $M$ dictates the memory bank's capacity, $B$ represents the training batch size, $K$ denotes the count of human parts, and $C$ signifies the feature dimensions of the parts embedding. This PEMB undergoes dynamic updates throughout training, replacing the oldest parts embeddings with the latest ones at each batch iteration.
\vspace{-3mm}
\paragraph{Inference.}
Consistent with prior studies~\cite{hcga, bpbreid}, our approach exclusively relies on foreground and part embeddings to recover occluded pedestrians during inference. Concerning part embeddings, we solely calculate the distance between the two sides sharing the visible part. The determination of visibility is contingent upon whether the maximum predicted probability exceeds 40\%.

\subsection{Optimization}
The overall objective used to optimize the DROP framework during the training stage is formulated as follows:
\begin{equation}
    \mathcal{L} = \mathcal{L}_{reid} + \mathcal{L}_{pct}+\lambda\mathcal{L}_{hp},
\end{equation}
where $\mathcal{L}_{reid}$ represents the cross-entropy loss incorporating label smoothing~\cite{labels} and the BNNeck trick~\cite{bot}, while $\mathcal{L}_{pct}$ and $\lambda\mathcal{L}_{hp}$ denote the part-aware compactness loss and spatial-smoothed parsing loss. The variable $\lambda$, set empirically to 0.4, governs the contribution of human parsing. Additionally, $\mathcal{L}_{reid}$ drives the optimization of DROP in predicting pedestrian image identity through global, foreground, and parts embedding. 
\vspace{-3mm}
\paragraph{Part-aware compactness triplet loss.}
Unlike the standard batch hard triplet loss~\cite{triple} that calculates distances between two pedestrians, our method computes distances among human parts utilizing the PEMB established during training. Since the inference process relies on shared visibility-based part-to-part matching, our focus lies in determining the distances between parts embeddings. However, due to potential occlusions leading to parts being excluded and resulting in inadequate part samples, we create and maintain a PEMB throughout the training phase, detailed in Section~\ref{sec:reidBranch}. Leveraging PEMB, we initially compute a pairwise distance matrix $\mathcal{M}_{parts}$ sized $[K, M\times B, M\times B]$ for $K$ parts embeddings $E_k$.
\begin{equation}
    \mathcal{M}_{parts} = dist(E^m_k, E^n_k)|(E^m, E^n)\in PEMB,
\end{equation}
where $dist$ denotes the Euclidean distance. Subsequently, we calculate the pairwise distance matrix for pedestrians, sized $[M\times B, M\times B]$, by amalgamating the part-based distances. Finally, the standard batch-hard triplet loss is computed utilizing the generated pairwise distance matrix.
\begin{equation}
    \mathcal{L}_{pct} = \left[Avg(\mathcal{M}^{ap}_{parts})-Avg(\mathcal{M}^{an}_{parts})+\alpha\right]_{+},
\end{equation}
where the distances from the anchor sample to the hardest positive and negative samples in PEMB are denoted by $\mathcal{M}^{ap}_{parts}$ and $\mathcal{M}^{an}_{parts}$ respectively, $Avg$ is the averaging operation and $\alpha$ is the triplet loss margin. The PCT loss optimizes the average distances among corresponding parts stored in PEMB. By ensuring ample negative samples for each human part, even amid occlusion, our PCT fosters the learning of robust and condensed features. This strategy aids in mitigating the impact of both occluded and non-discriminative local features.

\paragraph{Spatial-smoothed parsing loss.}
Given the imprecise nature of the rough human parsing results derived from predictions, particularly when relying on additional semantic models, we incorporate label smoothing~\cite{RethingReIDwithConf, labels} into the pixel-level cross-entropy loss. Additionally, aiming for spatially smoothed predicted outcomes, we introduce a straightforward spatial smoothing regularization term, formulated as follows:
\begin{equation}
\begin{aligned}
& \mathcal{L}_{hp}=\sum_{k=0}^K \sum_{h=0}^{H-1} \sum_{w=0}^{W-1} -q_k \cdot \log \left(hp_k(h, w)\right)  
\\ & +\gamma(\left\|hp_k(h+1, w)-hp_k(h, w)\right\|_1
\\ & +\left\|hp_k(h, w+1)-hp_k(h, w)\right\|_1),  
\\ & \text { with } q_k= \begin{cases}1-\frac{B-1}{B} \varepsilon & \text { if } \mathcal{H}(h, w)=k \\
\frac{\varepsilon}{B} & \text {otherwise, }\end{cases}
\end{aligned}
\end{equation}
where the first term is pixel-level cross-entropy loss with label smoothing, $\varepsilon$ is the label smoothing regularization rate, the second term is spatial smoothing, $\gamma$ is used to control the spatial smoothing contribution and is empirically set to 0.5, and $hp_k(h,w)$ is the prediction probability for part $k$ at spatial location $(h,w)$.

\section{Experiments}
\subsection{Experiments setup}
\paragraph{Dataset and Evaluation Metric.}

\begin{table}[]
\centering
\small
\caption{Performance comparison with state-of-the-arts on Occluded-Duke and P-DukeMTMC (\%). The first and second best results are labeled in \textbf{bold} and in \underline{underlined}.* means the results are reproduced with image size $256\times 128$.}
\label{tab:occulded_dataset}
\begin{tabular}{l|cc|cc}
\toprule
\multirow{2}{0pt}{Methods} & \multicolumn{2}{c|}{Occluded-Duke} & \multicolumn{2}{c}{P-DukeMTMC} \\
\cline{2-5}
                    &Rank-1   &mAP    &Rank-1   &mAP    \\
\hline
PCB~\cite{partlevel-pcb}					&42.6		&33.7		&79.4	&63.9		\\
DSR~\cite{DSR}         &40.8   &30.4   &-   &-   \\
SFR~\cite{SFR}      &42.3   &32.0     &-      &-      \\
\hline
PVPM~\cite{kepoint-pvpm}         &47.0     &37.7   &85.1   &69.9   \\
PGFA~\cite{keypoint-pgfa}        &51.4   &37.3   &85.7      &72.4      \\
HOReID~\cite{keypoint-honet}        &55.1   &43.8   &72.3  &62.9   \\
ISP~\cite{pixellevel-isp}        &62.8   &52.3   &89.0      &74.7      \\
QPM~\cite{QPM}        &66.7  &53.3  &90.7 &75.3    \\
MoS~\cite{MOS}        &66.6  &55.1  &- &-     \\
SSGR~\cite{SSGR}   & 69.0 & 57.2 & - & - \\
HCGA~\cite{hcga} &70.2 & 57.5 & - & -   \\
BPBreID*~\cite{bpbreid} & \underline{73.9} &62.0  & \underline{92.8} & \underline{83.1} \\
\hline
IGOAS~\cite{IGOAS}        &60.1  &49.4  &86.4 &75.0     \\
CAAO$_{ViT}$~\cite{caao}         &68.5 &59.5  &92.5  &81.4  \\
FCFormer$_{ViT}$~\cite{FCFormer} & 73.0 & \underline{63.1} & 92.4 & 82.5 \\
\hline
PAT$_{ViT}$~\cite{PAT}        &64.5  &53.6  &- &-     \\
TransReID$_{ViT}$~\cite{TransReID} &66.4		&59.2		&-	&-	\\	
FED$_{ViT}$~\cite{FED}        &68.1 &56.4  &- &-     \\
SAP$_{ViT}$~\cite{SAP}  & 70.0 & 62.2 & - & - \\
\hline
\textbf{DROP(Ours)}  & \textbf{76.8} &\textbf{63.3}  & \textbf{93.8} &\textbf{83.4}  \\

\bottomrule
\end{tabular}
\vspace{-3mm}
\end{table}

We evaluate our model DROP on three occluded datasets Occluded-Duke~\cite{PGFA}, Occluded-ReID~\cite{AFPB} and P-DukeMTMC~\cite{AFPB} and two holistic datasets Market-1501~\cite{market} and DukeMTMC-reID~\cite{dukemtmc}. Half of Occluded-REID is used for training and the remaining half for testing. 
Following most works in person ReID, the Cumulative Matching Characteristic curves (CMC) at Rank-1 and Rank-5 and the mean average precision (mAP) are used in this paper to evaluate the performance of different person ReID methods. All experiments are implemented on two NVIDIA RTX 3090 GPUs and in the single query setting without re-ranking~\cite{zhong2017re-ranking}. 
\vspace{-3mm}
\paragraph{Implementation and Training Details.}
The DROP framework is implemented based on torchreid~\cite{torchreid} built by Pytorch~\cite{pytorch}. All images of the training set are resized to $256\times 128$ and augmented with random erasing~\cite{RandomErasing}, horizontal flipping, random cropping, and padding 10 pixels. All parameters are trained for 120 epochs with the Adam optimizer. The learning rate is 3.5e-4 and decays to 0.1 at 40 and 70 epochs. The batch size is 64 and the size of the PEMB is 4. The label smoothing regularization rate $\varepsilon$ is set to 0.1 and the triplet loss margin $\alpha$ is set to 0.3. The human parsing labels $\mathcal{H}$ are generated using the 17-part confidence and 19-part affinity fields produced by the PifPaf~\cite{PifPaf} pose estimation model. Following~\cite{bpbreid}, we split heatmaps into $K$ group. For occluded and holistic datasets, $K$ is set to 8 and 5, respectively.
Some existing approaches use SCHP~\cite{Self-Corr} or weakly supervised methods (e.g., cascade clustering~\cite{pixellevel-isp} or human co-parsing networks~\cite{hcga}) to generate coarse human parsed labels, which only yield worse performance compared to PifPaf. The possible reason is that PifPaf provides consistent predictions with few false negatives on a wide range of image resolutions~\cite{bpbreid}.
An ablation study of $K$ is in the Appendix. 

\subsection{Comparisons with State-of-the-arts}

\begin{table}[t]
\centering
\small
\caption{Comparison with state-of-the-art methods on Occluded-REID datasets (\%).}
\label{tab:OREID}
\begin{tabular}{lc|cc}
\toprule
Methods & References & Rank-1  & Rank-5          \\
\hline
SVDNet~\cite{svdnet}                 & ICCV17                      & 63.1            & 85.1            \\
 MLFN~\cite{MLF}                   & CVPR18                      & 64.7            & 87.7            \\
PCB~\cite{partlevel-pcb}                    & ECCV18                      & 66.6            & 89.2            \\
 AFPB~\cite{pixellevel-icme}                   & ICME18                     & 68.1            & 88.3            \\
Teacher-S~\cite{Teacher-S}              & Arxiv18                     & 73.7            & 92.9 \\
 REDA~\cite{RandomErasing}                   & AAAI20                      & 65.8            & 87.9            \\       
ISP~\cite{pixellevel-isp}                    & ECCV20                     & 86.2               & 95.4     \\ 
IGOAS~\cite{IGOAS} & TIP21 & 81.1 & 91.6  \\
HCGA~\cite{hcga}    & TIP23 & 88.0  & 96.0  \\
BPBreID~\cite{bpbreid} & WACV23 & \underline{93.8} & \underline{98.0} \\
\hline
\textbf{DROP(Ours)}  & & \textbf{94.2} &\textbf{98.2}\\
\bottomrule

\end{tabular}
\vspace{-3mm}
\end{table}

\paragraph{Results on Occluded Datasets.} As shown in Table~\ref{tab:occulded_dataset}, we compare our method with 
5 holistic person ReID methods: SVDNet~\cite{svdnet}, DSR~\cite{DSR}, PCB~\cite{partlevel-pcb}, SFR~\cite{SFR}, MLFN~\cite{MLF}, 
10 state-of-the-art (SOTA) person occluded ReID methods: AFPB~\cite{AFPB}, Teacher-S~\cite{Teacher-S}, PGFA \cite{PGFA}, HOReID~\cite{keypoint-honet}, ISP \cite{pixellevel-isp}, QFM~\cite{QPM}, MOS~\cite{MOS}, SSGR~\cite{SSGR}, HCGA~\cite{hcga}, BPBreID~\cite{bpbreid}, 3 data augmentation based methods: REDA~\cite{RandomErasing}, IGOAS~\cite{IGOAS}, CAAO~\cite{caao}, FCFromer~\cite{FCFormer}
and 4 transformer-based ReID method: PATrans~\cite{PAT}, TransReID~\cite{TransReID}, FED~\cite{FED}, and SAP~\cite{SAP}. For the Occluded-Duke and P-DukeMTMC datasets, the occluded ReID methods are about 20\% higher than the holistic ReID methods in Rank-1 and mAP. Compared to CNN-based methods, Vision Transform(ViT)--based methods usually achieve better results on mAP, and in particular, FCFormer achieves the second-best mAP performance on Occluded-Duke.
Compared with the second-best CNN-based method BPBReID, DROP improved by 2.9\% in Rank-1 and 1.3\% in mAP. 
Compared with the ViT-based approach, DROP achieved similar mAP and 3.8\% Rank-1 improvement.

For a fair comparison of the Occluded-ReID dataset, we do not list the performance of FCFormer and CAAO in Table~\ref{tab:OREID}. This is because those methods use a different dataset division method from AFPB, which proposes the Occluded-REID dataset. 
Similarly, our method achieve better performance in Rank-1 and Rank-5 compared with the second-best method BPBreID. 

\begin{table}[]
\centering
\small
\caption{Performance comparison with state-of-the-art methods on Market-1501 and DukeMTMC-reID datasets (\%). The first and second best results are labeled in \textbf{bold} and in \underline{underlined}.
}
\label{tab:holistic_dataset}
\begin{tabular}{l|cc|cc}
\toprule
\multirow{2}{0pt}{Methods}  & \multicolumn{2}{c|}{Market-1501} & \multicolumn{2}{c}{DukeMTMC} \\
\cline{2-5}
                  &Rank-1   &mAP    &Rank-1   &mAP    \\
\hline
PCB+RPP~\cite{partlevel-pcb}      &92.3   &77.4   &81.8   &66.1   \\
MGN~\cite{mgn}     &95.7   &86.9   &88.7   &78.4   \\
MHN-6~\cite{attention1}      &95.1   &85.0   &89.1   &77.2   \\
\hline
SPReID~\cite{pixellevel-hsp}    &92.5   &81.3   &84.4   &71.0   \\

$P^{2}$ Net~\cite{pixellevel-p2net}  &95.2   &85.6   &86.5   &73.1   \\
PGFA~\cite{PGFA}      &91.2   &76.8   &82.6   &65.5   \\
HOReID~\cite{keypoint-honet}      &94.2   &84.9   &86.9   &75.6   \\
FPR~\cite{fpr}     &95.4   &86.6   &88.6   &78.4   \\
MoS~\cite{MOS}        &95.4  &89.0  &90.6 &80.2     \\
ISP~\cite{pixellevel-isp} &95.3 &88.6 & 89.6 & 80.0 \\ 
MPN~\cite{MPN} & \textbf{96.3}  & 89.4 & 91.5 & 82.0 \\
SSGR~\cite{SSGR} & \underline{96.1} & 89.3 & 91.1 & 81.3 \\
HCGA~\cite{hcga}    & 95.2 & 88.4 & 90.0 & 80.7  \\
BPBreID*~\cite{bpbreid} & 95.3  &88.8   & \underline{91.7} &  \underline{83.5}   \\
\hline
IGOAS~\cite{IGOAS}        &93.4  &84.1  &86.9 &75.1     \\
CAAO$_{ViT}$~\cite{caao}          &95.3		&88.0	&89.8	&80.9  \\
FCFormer$_{ViT}$~\cite{FCFormer} & 95.0 & 86.8 & 89.7 & 78.8 \\
\hline
PAT$_{ViT}$~\cite{PAT}     &95.4		&88.0	&88.8	&78.2 \\
FED$_{ViT}$~\cite{FED}        &95.0  &86.3  &89.4 &78.0     \\
TransReID$_{ViT}$~\cite{TransReID}      &95.0   &88.8   &90.4   &81.8   \\
SAP$_{ViT}$~\cite{SAP} & 96.0 & \underline{90.5} & - & - \\ 
NFormer~\cite{NFormer} & 94.7 & \textbf{91.1} & 89.4 & \underline{83.5} \\
\hline
\textbf{DROP}(\textit{Ours}) & 95.6	& 89.5	& \textbf{92.8}	& \textbf{84.3} \\
\bottomrule
\end{tabular}
\vspace{-3mm}
\end{table}

\vspace{-3mm}
\paragraph{Results on Holistic Datasets.}
As shown in Table~\ref{tab:holistic_dataset}, we compare the proposed method with 
3 part-level alignment-based methods: PCB+RPP~\cite{partlevel-pcb}, MGN~\cite{mgn}, MHN-6~\cite{attention1},   
11 alignment-based methods: SPReID~\cite{pixellevel-hsp}, $P^2$-Net~\cite{pixellevel-p2net}, PGFA~\cite{PGFA},
HOReID~\cite{keypoint-honet},  FPR~\cite{fpr}, ISP~\cite{pixellevel-isp}, MPN~\cite{MPN}, SSGR~\cite{SSGR}, HCGA~\cite{hcga}, BPBreID~\cite{bpbreid}, 
3 data augmentation based methods: REDA~\cite{RandomErasing}, IGOAS~\cite{IGOAS}, CAAO~\cite{caao}, FCFromer~\cite{FCFormer}
and 4 transformer-based ReID method: PATrans~\cite{PAT}, TransReID~\cite{TransReID}, FED~\cite{FED}, SAP~\cite{SAP}, and NFormer~\cite{NFormer}.
Methods designed for the occlusion problem usually do not achieve optimal performance on holistic ReID datasets. For example, HOReID or FCFormer do not perform as well as the generic TransReID. The holistic ReID method MPN uses two additional types of information, human paring~\cite{liao2016understand} and human segmentation~\cite{pixellevel-maskg}, to achieve the best Rank-1 on Market-1501. Compared with the state-of-the-art methods in different directions, our method still achieves comparable performance on Market-1501 and first-best Rank-1 and mAP on the DukeMTMC.

\subsection{Ablation Study}

\begin{table}[!t]
\centering
\small
\caption{Ablation study for the main components of DROP on the Occluded-Duke (\%). ``Decouple" is the decoupled branches, ``PPF" is Pedestrian Position-aware Features, ``PCT" is Part-aware Compactness Triplet loss, and ``SS" is the spatial smoothing term.}
\label{tab:ablation_com}
\begin{tabular}{ccccc|cc}
\toprule
Baseline & Decouple & PPF & PCT  & SS & R-1  & mAP         \\ \hline
\checkmark &  &  &  &            & 57.8      & 49.6  \\    
\checkmark & \checkmark  &  &          &   &   73.5  & 61.3 \\    
\checkmark & \checkmark  & \checkmark &    &   &74.6  & 61.7\\ 
\checkmark & \checkmark  &  \checkmark& \checkmark      &   &  76.2  & 62.8 \\ 
\checkmark & \checkmark  &  \checkmark&       &  \checkmark &  75.2  & 62.2 \\
\checkmark & \checkmark  &  \checkmark&  \checkmark   & \checkmark & \textbf{76.8}  & \textbf{63.3} \\ 
\bottomrule
\end{tabular}
\vspace{-2mm}
\end{table}

\paragraph{Components of DROP.}
As shown in Table~\ref{tab:ablation_com}, we adopt HRNet-W32~\cite{HRNet} as a baseline and build DROP on top of it. 
First, unlike previous approaches, our focus on solving task-specific features brings huge performance gains. Compared with the couple branches, our method only slightly increases the computational cost, demonstrating the good efficiency of our design. We further visualize the classification and parsing loss and accuracy when training Baseline with and without Decouple branches in the Appendix. Decoupled branches can accelerate the training and contribute to better convergence.
Next, On the parsing branch, we add pedestrian location-aware features, which also bring good performance improvement. 
Finally, we analyze the improvements of Drop in terms of loss. First, we add a spatial smoothing regularity term to the regular segmentation loss. We expect this regular term to implicitly constrain the parsing results, i.e., to be locally spatially similar for each human part. Second, for parts embeddings, we propose generalized PCT to enhance model learning occluded body parts and non-discriminative local appearance.

\vspace{-3mm}
\paragraph{Validation of the generality of PCT loss.}
In this section, we verify that the proposed PCT loss is not only useful in DROP but is valid for both part-based methods. We compare part-level Hard Mining Center-Triplet (HCT) Loss~\cite{htt_loss}, part-average triplet loss~\cite{bpbreid} with our PCT loss on DROP and popular part-based ReID method PCB~\cite{partlevel-pcb}. 
The first two only do not take into account the lack of negative samples due to the lack of samples in the case of occlusion, the loss is optimized very quickly but does not learn compact parts embeddings. In contrast, we utilize PEMB to preserve a sufficiently large number of negative samples for the model to learn, and therefore achieve the best performance on the different methods.

\begin{table}[!t]
\centering
\small
\caption{Analysis of different triplet loss on Occlude-Duke (\%). PCB* is reproduced with our framework without parsing branch.}
\label{tab:backbone}
\begin{tabular}{l|cc|cc} 
\toprule
\multirow{2}{*}{Loss} & \multicolumn{2}{c}{DROP} & \multicolumn{2}{c}{PCB*} \\ \cline{2-5} 
                                           & R-1           & mAP            & R-1            & mAP             \\ \hline
Part-level HCT Loss   & 73.8         & 61.3          & 61.8             & 50.4            \\
Part-Average Triplet Loss   & 74.6             & 61.7           & 62.2           & 50.5           \\ 
PCT Loss & \textbf{76.2}   & \textbf{62.8}  & \textbf{63.2}  & \textbf{52.3}            \\ 
\bottomrule
\end{tabular}
\vspace{-2mm}
\end{table}

\vspace{-3mm}
\paragraph{Affect of different output embeddings.} As shown in Table~\ref{tab:embedding}, we study the discriminative ability of the holistic and human parts embeddings. First, for human parts embeddings, the upper body generally achieves better performance, because occlusion often occurs in the lower body. Second, the best performance was achieved using parts embedding alone, which demonstrates the effectiveness of part-to-part matches. Finally, although $G+F+P$ achieved the best mAP, balancing other metrics, we used $F+P$ for retrieval in all datasets.

\begin{table}[!t]
\centering
\small
\caption{Perfomance comparison for the global, foreground, and each human parts embeddings on Occluded-Duke (\%). $G$, $F$, and $P$ represent the global, foreground, and human parts embeddings. $k=\{1, \cdots, 8\}$ represents head, torso, right arm, left arm, right leg, left leg, right foot, left foot.}
\label{tab:embedding}
\begin{tabular}{l|cccc}
\toprule
Embeddings & mAP & Rank-1  & Rank-5  & Rank-10 \\\hline
$k=1$&  26.4  & 44.7    & 62.2 & 69.0      \\
$k=2$&  29.1 & 50.3    & \textbf{65.0}  & 70.3 \\
$k=3$&  29.2  & \textbf{50.5}   & 64.9    & \textbf{70.7}  \\       
$k=4$&   \textbf{30.5} & 48.6  & 62.0     & 67.7  \\ 
$k=5+6$& 17.1 & 28.8 & 46.3 & 55.0 \\
$k=7+8$& 7.0&  12.2 & 20.3 & 25.3  \\ \hline
$G$&  54.9  &  64.0   & 77.2     &  82.3  \\
$F$&  57.3 & 69.2  & 82.8   & 86.7       \\
$P$&  \textbf{61.3} & \textbf{75.2}   & \textbf{86.5}  & \textbf{89.7}    \\ \hline
$G+F$& 58.2& 68.1 & 81.3 & 85.8 \\
$F+P$& 63.3  & \textbf{76.8} & \textbf{87.2} & \textbf{92.7} \\
$G+F+P$& \textbf{63.6} & 75.8 & 86.8 & 90.2 \\
\bottomrule
\end{tabular}
\vspace{-2mm}
\end{table}

\vspace{-3mm}
\paragraph{Affect of different backbones.}
We analyze the impact of different backbones. As demonstrated in Table~\ref{tab:backbone}, we compare HRNet-W32~\cite{HRNet} with ResNet50~\cite{ResNet} and ResNet50-IBN~\cite{ibn-net}. For ResNet50 and ResNet50-IBN, we directly use the upsampling layer to linearly interpolate the $16\times8$ feature map to $64\times32$, in keeping with HRNet. For different backbones, HRNet achieves the best performance. That is because the primitive resolution of the feature maps may be the main factor affecting the performance~\cite{pixellevel-isp, HRNet} for multi-tasking frameworks with segmentation as an auxiliary task. More importantly, our approach outperforms existing multi-tasking framework approaches with the same backbone.

\begin{table}[!t]
\centering
\small
\caption{Analysis of the backbone on Occluded-Duke (\%). "Param" denotes the parameters of the Backbone.* means the results are reproduced with image size $256\times 128$.}
\label{tab:backbone}
\begin{tabular}{lll|cccc} 
\toprule
Backbone                       & Param                  & Methods & Rank-1 & mAP  \\
\hline
\multirow{3}{*}{ResNet-50}     & \multirow{3}{*}{28.1M} & HCGA    & 61.0   & 45.9 \\
                               &                        & BPBreID* & 66.4   & 52.7 \\
                               &                        & DROP    & \textbf{69.3}   & \textbf{54.0} \\
\hline
\multirow{2}{*}{ResNet-50-IBN} & \multirow{2}{*}{28.1M} & BPBreID* & 70.9   & 56.6 \\
                               &                        & DROP    & \textbf{72.4} & \textbf{58.0} \\
\hline
\multirow{4}{*}{HRNet-W32}     & \multirow{4}{*}{28.5M} & ISP & 62.8& 52.3 \\
                               &                &HCGA    & 70.2   & 57.5 \\
                               &                        & BPBreID* & 73.9   & 62.0 \\
                               &                   & DROP    & \textbf{76.8}   & \textbf{63.3}\\
\bottomrule
\end{tabular}
\vspace{-2mm}
\end{table}

\subsection{Qualitative Results}
\begin{figure}[!t]
\centering
\includegraphics[width=\linewidth]{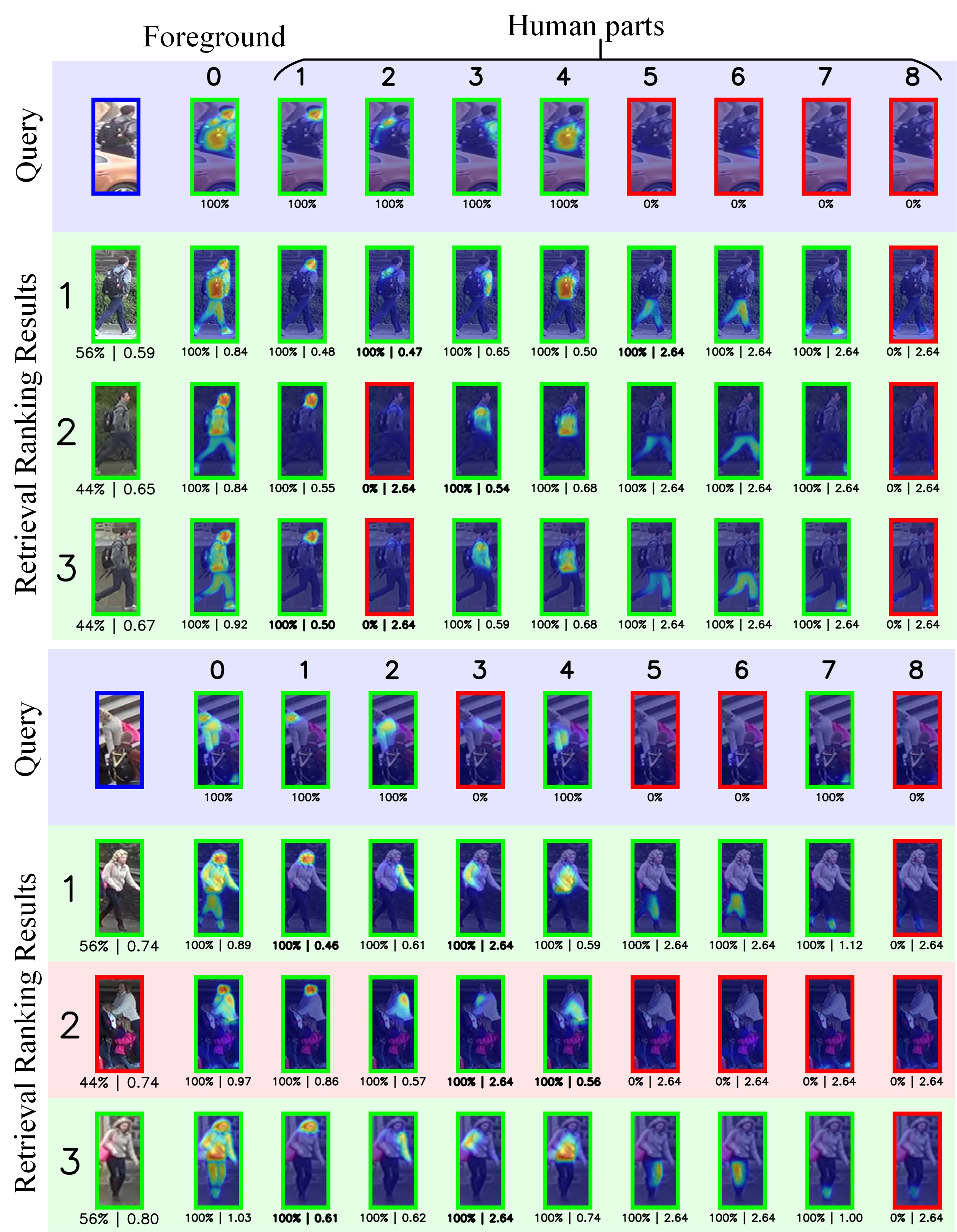}
\caption{Qualitative results of DROP. The blue box represents the query, the green box in the first column indicates a successful retrieval, and the red box indicates a failed retrieval. The green boxes in the next 9 columns indicate that the human part is partitioned, while the red boxes indicate that there is no corresponding human part.
}
\label{fig:vis}
\vspace{-3mm}
\end{figure}

We present visualization results of DROP for two distinct occlusion scenarios in Fig.~\ref{fig:vis}. In the instance where pedestrians encounter occlusion due to objects in the scene, our model, guided by the outcomes of the human parsing branch, selectively focuses solely on the pedestrians. On the other hand, when pedestrians occlude each other, we classify the occluding pedestrians as background, utilizing positional information for this determination. Additional results can be found in the Appendix.

\section{Conclusion}
In this paper, we first analyze the essential reasons why present multitasking frameworks incorporating human parsing perform poorly. Based on this, we propose to decouple person re-identification from human parsing and present two branches to learn task-specific features. For the human position-aware parsing branch, we take one-dimensional height information as input and let the network learn pedestrian position embedding. For the parsing-guided ReID branch, we update a parts embedding memory bank during training for part-aware compactness triplet loss learning. The effectiveness of our method is demonstrated on three occluded datasets and two holistic datasets.

{
    \small
    \bibliographystyle{ieeenat_fullname}
    \bibliography{main}
}

\clearpage
\setcounter{page}{1}
\maketitlesupplementary

\section{More Related Work}
\paragraph{Vision Transformer-based Person Re-identification}
Compared to existing CNN-based methods, transformer-based approaches demonstrate superior resilience to occlusion. He~\textit{et al.}~\cite{TransReID} were the pioneers in harnessing the pure Transformer for ReID tasks, presenting the Transformer-based Object Re-identification (TransReID) method. TransReID incorporates side information embedding for encoding various contextual cues and introduces the jigsaw patches module to implement the stripe-based concept.
Li~\textit{et al.}~\cite{PAT} pioneer the exploration of a transformer encoder-decoder structure for Occluded ReID. They introduce the Part-Aware Transformer (PATrans) for learning part prototypes, incorporating part diversity and discriminability to enhance robust human part discovery.
Jia \textit{et al.} \cite{learning_dis_transformer} present a disentangled representation learning network (DRL-Net) designed to address occluded Re-ID challenges without the need for precise person image alignment.

\section{Structure of WAMP}
Similar to GWAP~\cite{GWAP}, we initially acquire the parsing outcomes alongside ReID features for element-wise multiplication. Subsequently, we employ two distinct pooling methods to condense the features. Upon aggregating these compressed features, a fully connected layer is utilized to reduce the dimensionality of the resulting features. WAMP provides a slight performance boost compared to GWAP.

\label{sec:intro}
\begin{figure}[!t]
\centering
\includegraphics[width=\linewidth]{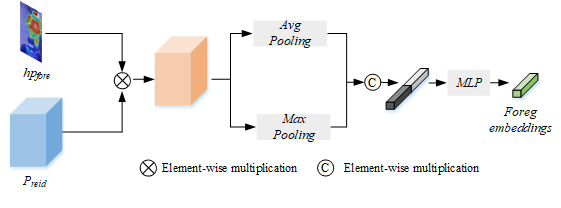}
\caption{The structure of Weight Average and Max pooling.}
\label{diff}
\vspace{-3mm}
\end{figure}

\section{More Experiment Results}

\subsection{More Ablation study}
\label{sec:ablation_K}

\begin{table}[t]
\centering
\caption{Performance comparison for the different number of body parts on Occluded-Duke (\%). }
\label{tab:num_K}
\begin{tabular}{l|cccc}
\toprule
Embeddings & mAP & Rank-1  & Rank-5  & Rank-10 \\\hline
$K=3$&  57.2  & 69.9   & 82.9 & 86.9    \\
$K=4$&  63.0 & 73.7    & 85.3  & 89.0 \\
$K=5$&  63.6 & 75.2  & 86.4   & 89.4 \\       
$K=6$&   \textbf{63.4} & 76.3  & 86.9    & 89.6  \\ 
$K=7$& 62.5 & 74.0 & 85.5 & 89.3 \\
$K=8$ & 63.3  & \textbf{76.8} & \textbf{87.2} & \textbf{92.7} \\
\bottomrule
\end{tabular}
\vspace{-2mm}
\end{table}

\paragraph{Ablation study on the number of body parts $K$}
In this section, we explore the impact of the number of body parts $K$ predicted by the human position-aware parsing branch. The effective training of the human position-aware parsing branch requires the utilization of pre-generated human parsing labels, which are 2D human semantic segmentation maps assigning integer values from 0 to $K$ to each pixel. Here, 0 represents the background label, while values between 1 and $K$ denote the labels of the $K$ body regions. Table~\ref{tab:num_K} presents the performance rankings for various $K$ values on the Occluded-Duke dataset. The optimal performance is observed at $K = 8$. However, exceeding this value leads to an escalation in model parameters, surpassing the maximum GPU memory capacity of our device. Conversely, performance decreases when $K$ is below 8.

\paragraph{1D Position encoding VS. 2D position encoding}
In our analysis, we examine the impact of various positional coding schemes. While the 1D approach incorporates solely height information, the 2D method incorporates both horizontal and vertical data. However, the distinction between top and bottom is clearer compared to that between left and right. For instance, distinguishing between left and right hands can be challenging when considering the object's front and back perspectives. 
\begin{table}
\centering
\caption{Performance comparison for the 1D position encoding and 2D position on Occluded-Duke (\%). }
\begin{tabular}{l|cc}
\toprule
Methods & mAP & Rank-1 \\\hline
Decoupled Branches &  61.3  & 73.5     \\
+ 1D Position Encoding &  \textbf{61.7} & \textbf{74.6}    \\
+ 2D Position Encoding &  61.6 & 73.2   \\       
\bottomrule
\end{tabular}
\label{tab:1Dvs2D}
\end{table}

\subsection{More Qualitative Results}

\begin{figure*}[!t]
\centering
\includegraphics[width=\linewidth]{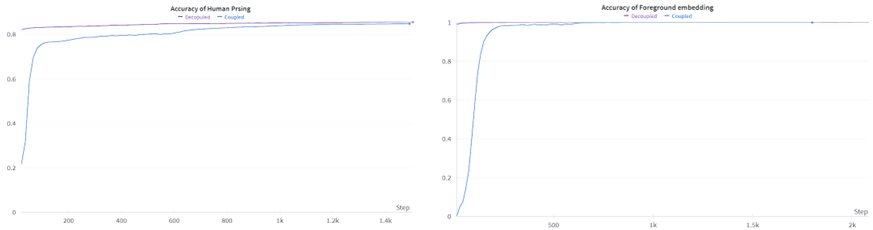}
\caption{The accuracy in the training processing.\textbf{Left:} the accuracy of human parsing. \textbf{Right:} the accuracy of foreground embedding.}
\label{diff}
\vspace{-3mm}
\end{figure*}

\begin{figure*}[!t]
\centering
\includegraphics[width=\linewidth]{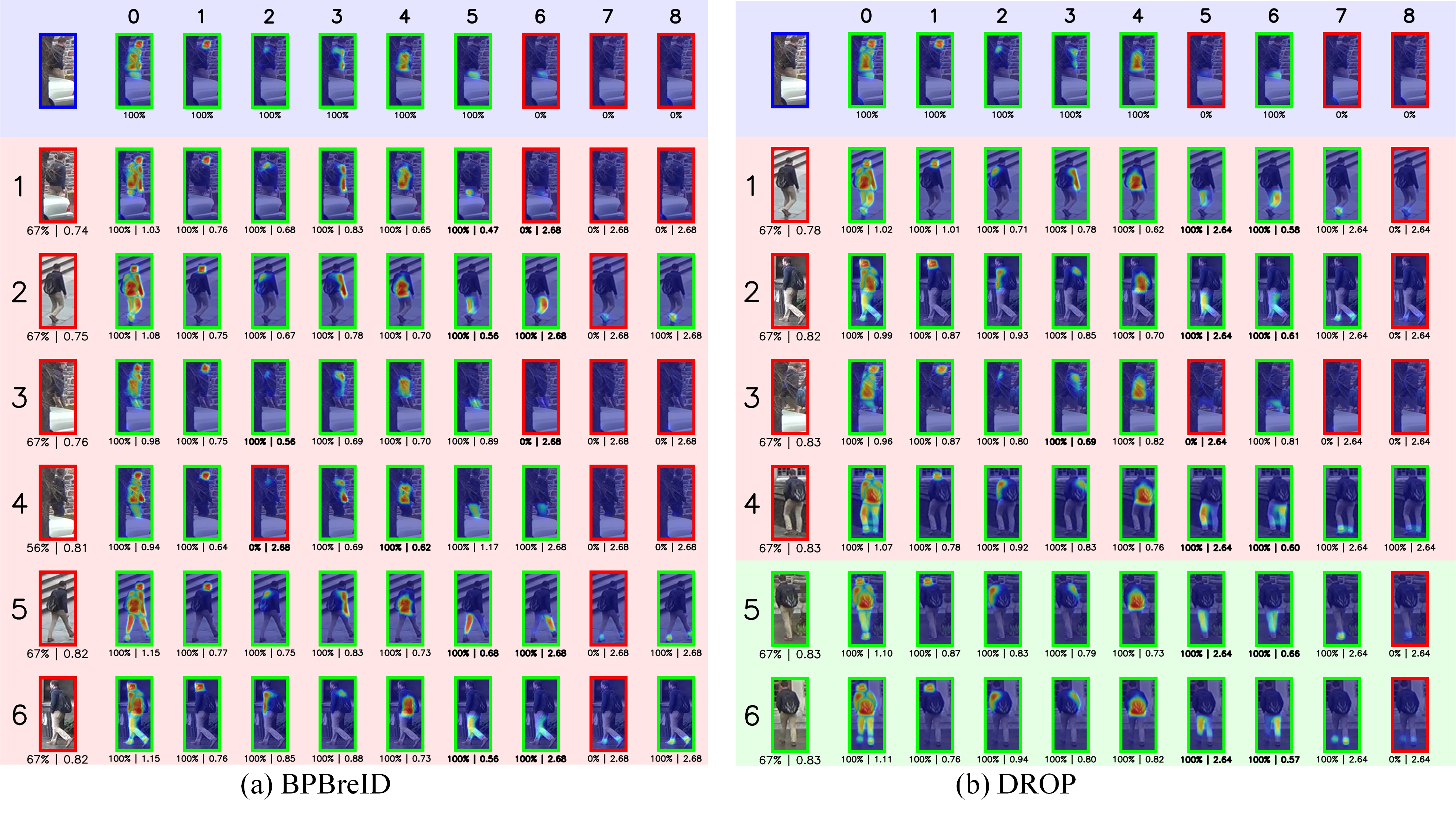}
\caption{Comparison of the ranking performance of our model DROP with BPBreID.}
\label{diff}
\vspace{-3mm}
\end{figure*}


\end{document}